\documentclass{article}

\usepackage[preprint]{neurips_2026}

\usepackage[utf8]{inputenc} 
\usepackage[T1]{fontenc}    
\usepackage{hyperref}       
\usepackage{url}            
\usepackage{booktabs}       
\usepackage{amsfonts}       
\usepackage{nicefrac}       
\usepackage{microtype}      
\usepackage{xcolor}         
\usepackage{multirow}    
\usepackage{graphicx}    
\usepackage{wrapfig}      
\usepackage{caption}      
\usepackage{amsmath}      
\title{TRUE: A Trustworthy Unified Explanation Framework for Large Language Model Reasoning}

\author{
Yujiao Yang \\
Dalian University of Technology \\
Dalian, China \\
\texttt{yjyang@mail.dlut.edu.cn}
}

\begin{document}

\maketitle

\begin{abstract}
Large language models (LLMs) have demonstrated strong capabilities in complex reasoning tasks, yet their decision-making processes remain difficult to interpret. Existing explanation methods often lack trustworthy structural insight and are limited to single-instance analysis, failing to reveal reasoning stability and systematic failure mechanisms. To address these limitations, we propose the Trustworthy Unified Explanation Framework (TRUE), which integrates executable reasoning verification, feasible-region directed acyclic graph (DAG) modeling, and causal failure mode analysis. At the instance level, we redefine reasoning traces as executable process specifications and introduce blind execution verification to assess operational validity. At the local structural level, we construct feasible-region DAGs via structure-consistent perturbations, enabling explicit characterization of reasoning stability and the executable region in the local input space. At the class level, we introduce a causal failure mode analysis method that identifies recurring structural failure patterns and quantifies their causal influence using Shapley values. Extensive experiments across multiple reasoning benchmarks demonstrate that the proposed framework provides multi-level, verifiable explanations, including executable reasoning structures for individual instances, feasible-region representations for neighboring inputs, and interpretable failure modes with quantified importance at the class level. These results establish a unified and principled paradigm for improving the interpretability and reliability of LLM reasoning systems.
\end{abstract}

\section{Introduction}

Large language models (LLMs) have demonstrated remarkable capabilities on complex reasoning tasks spanning mathematical problem solving, logical inference, and knowledge-intensive decision-making. Understanding \emph{how} LLMs arrive at their decisions is crucial for revealing reasoning boundaries, building user trust, and enabling reliable deployment~\cite{lanham2023measuring, paul2024making, tutek2025measuring}. However, due to the scale and opacity of modern LLMs, traditional perturbation- and attention-based attribution methods fail to provide structurally meaningful explanations~\cite{agarwal2024faithfulness}. Prompting-based methods instead elicit step-by-step reasoning traces~\cite{cahlik2025reasoning}, yet prior work shows these traces often serve as post-hoc rationalizations rather than faithful reflections of the decision process~\cite{turpin2023language, barez2025chain}. Moreover, existing local explanation methods treat each instance in isolation, failing to expose structural patterns or systematic failure modes at the problem-class level.

To address these limitations, we propose the \textbf{Trustworthy Unified Explanation Framework (TRUE)}, a multi-level framework that elevates explanations from descriptive text to verifiable, executable, and structurally grounded reasoning representations. TRUE consists of three components: (i) \emph{executable explanations}, in which reasoning traces are redefined as process specifications and validated by an independent blind execution verifier that recovers the answer solely from the steps, establishing a direct link between explanation executability and reasoning correctness; (ii) \emph{local feasible-region DAGs}, constructed via structure-preserving perturbations, which characterize the stability and executability of reasoning across semantically similar inputs; and (iii) \emph{cluster-level failure mode analysis}, which aggregates counterfactual perturbations across semantically related samples and uses Shapley attribution to identify and quantify recurring structural failure modes at the class level.

We conduct extensive experiments on six representative reasoning benchmarks. The results show that executable explanations reliably capture the validity of reasoning structures at the instance level, feasible-region DAGs effectively model structural stability, and cluster-level analysis identifies interpretable and quantifiable systematic deficiencies. Together, these components constitute a unified paradigm for interpretable and verifiable LLM reasoning analysis.

\section{Related Works}

\begin{wrapfigure}{r}{0.5\linewidth}
\vspace{-10pt}
\centering
\includegraphics[height=0.4\textheight]{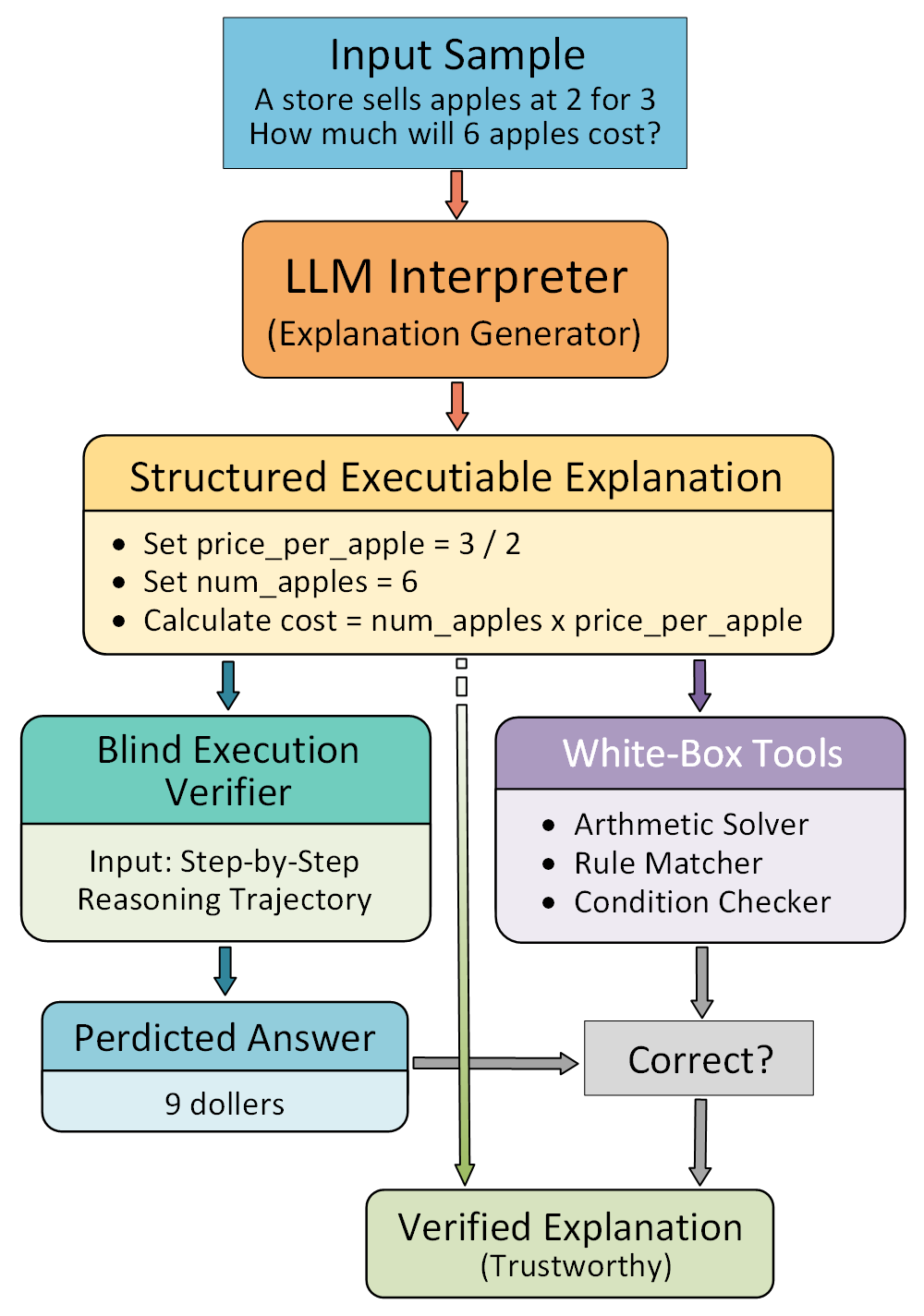}
\vspace{-3pt}
\caption{Generation and verification of executable explanations. Given an input instance, the LLM interpreter produces a structured executable explanation consisting of explicit reasoning steps. The explanation is executed by a blind verifier to obtain a predicted answer, while white-box tools independently verify correctness of each reasoning step.}
\vspace{-0.8cm}
\label{fig:1}
\end{wrapfigure}

\paragraph{Perturbation-Based Local Explanation.}
Perturbation-based methods estimate input importance by applying controlled modifications and observing output changes, forming one of the most widely used model-agnostic paradigms~\cite{ribeiro2016should}. For LLMs, tokens, context segments, or prompts are selectively removed or replaced to analyze their influence on reasoning behavior~\cite{turpin2023language, madani2025noiser, ortu2024competition}. Shapley-value methods provide a theoretically grounded framework by quantifying marginal contributions~\cite{lundberg2017unified}; recent extensions adapt them to LLM token- and context-level attribution~\cite{horovicz2024tokenshap, xiao2025tokenshapley, liu2024attribot, patel2025maxshapley}, as well as data attribution~\cite{tan2025understanding, naudot2025llmshap} for analyzing the influence of training data or in-context examples. Despite their rigor, these methods identify influential components in isolation without modeling the structural organization of reasoning, and therefore cannot verify executability, characterize stability across related inputs, or identify failure modes.

\paragraph{Prompting-Based Explanation.}
Prompting-based methods elicit intermediate reasoning steps as natural-language traces. Chain-of-Thought (CoT)~\cite{wei2022chain} and its extensions—Zero-Shot CoT~\cite{kojima2022large}, Plan-and-Solve~\cite{wang2023plan}, Self-Refine~\cite{madaan2023self}—produce coherent step-by-step reasoning. These approaches make the reasoning process more observable and provide a practical interface for analyzing model behavior. Subsequent studies analyze how such traces reflect model behavior via demonstration editing and contradictory mappings~\cite{han2023understanding, wei2023larger}, perturbation/attribution on CoT~\cite{wang2022self, chuang2025selfcite}, and counterfactual prompting~\cite{madaan2022language, zhou2023context}. However, faithfulness concerns persist: generated traces may act as post-hoc rationalizations uncorrelated with the true decision process~\cite{turpin2023language, lanham2023measuring}. Overall, prompting-based methods treat explanations as descriptive text without verifying whether they constitute complete and executable reasoning procedures, operate only at the single-instance level, and do not model shared reasoning structures or systematic failure modes, which motivates our framework.

\section{Methodology}

\subsection{Executable Explanations for Single-Instance Reasoning}

We begin by studying \emph{trustworthy explanations} at the single-instance level. Since CoT-style traces often act as post-hoc rationalizations rather than faithful decision records, we redefine explanations as \emph{executable reasoning specifications} whose validity can be verified by external execution (Figure~\ref{fig:1}).

\paragraph{Definition 1 (Executable Explanation).}
Given an input problem $x$, an executable explanation is defined as a structured sequence of reasoning steps
\begin{equation}
E = (e_1, e_2, \dots, e_T),
\end{equation}
where each step $e_t$ specifies an executable operation with its symbolic variables and a brief description. The explanation is a process specification, not a narrative, and must not reveal intermediate numerical values or the final answer. In practice, candidate explanations are generated via fixed prompting strategies (e.g., CoT, Plan-and-Solve) and uniformly treated as executable explanations.

\paragraph{Definition 2 (Trustworthy Explanation).}
$E$ is \emph{trustworthy} if an independent executor can recover the correct answer without access to the original problem $x$. We implement this via a \emph{blind execution verifier} $V$ that combines LLM-based step interpretation with white-box components (arithmetic solvers, rule-based matchers) to prevent external knowledge injection: the LLM only interprets the provided steps while white-box modules ensure operational correctness. The verifier receives only $E$ and produces
\begin{equation}
\hat{y} = V(E).
\end{equation}
If $\hat{y}$ matches the ground truth, the explanation encodes a complete and executable solution structure and is deemed trustworthy.

\begin{wrapfigure}{r}{0.5\linewidth}
\vspace{-10pt}
\centering
\includegraphics[height=0.4\textheight]{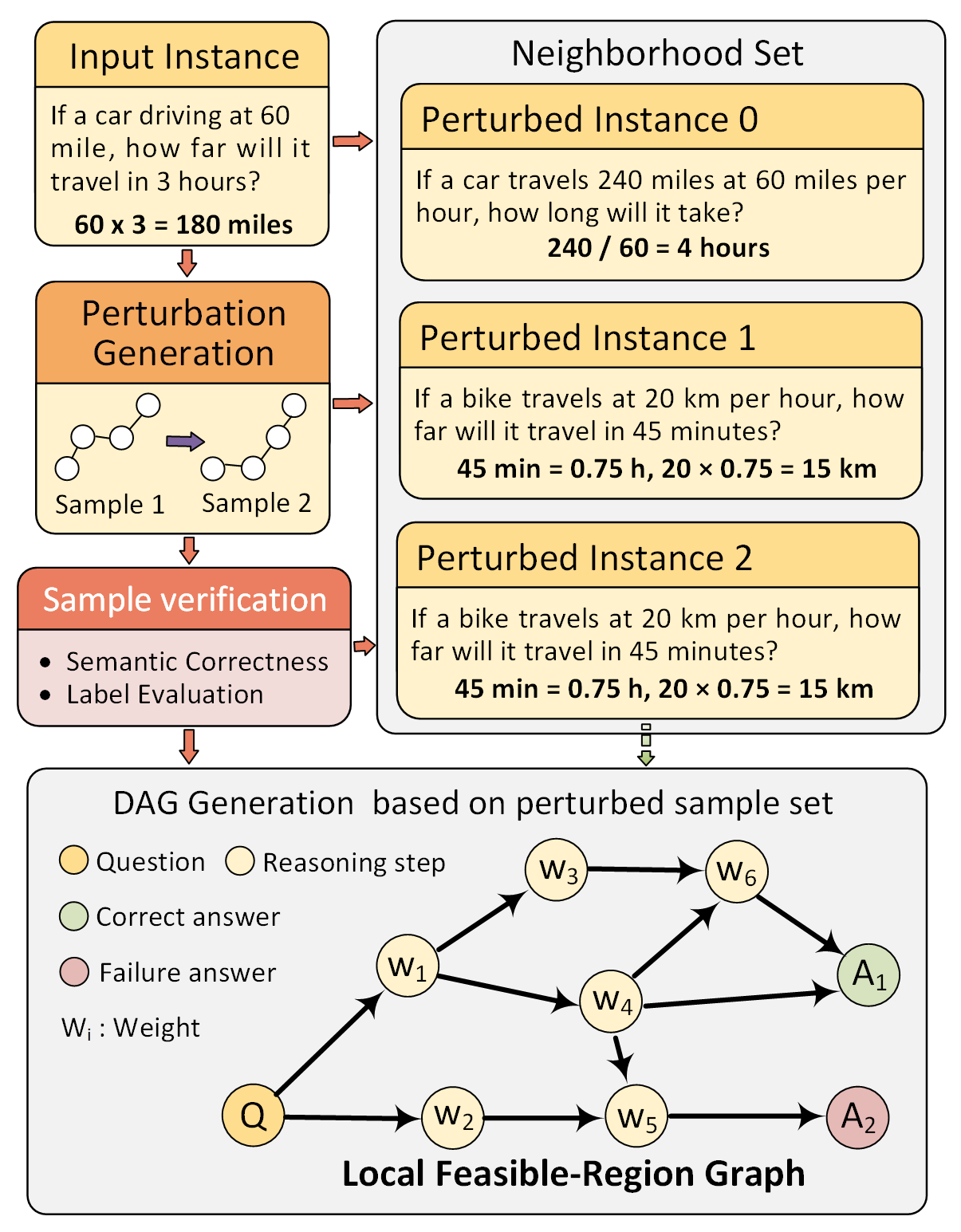}
\vspace{-12pt}
\caption{Construction of the local feasible-region graph via structure-preserving perturbations. Verified perturbed samples are aggregated into a directed acyclic graph, where nodes represent reasoning steps and edges encode their dependencies.}
\vspace{-1.2cm}
\label{fig:2}
\end{wrapfigure}

\subsection{Local Feasible-Region Modeling via Structure-Preserving Perturbations}

Executable explanations verify single-instance validity but do not characterize stability under nearby input variations. To assess whether the model preserves correct reasoning across semantically similar inputs, we propose a \emph{local feasible-region modeling} framework that characterizes the feasible reasoning structure in the input neighborhood (Figure~\ref{fig:2}).

\paragraph{Structure-Preserving Perturbation Generation.}
Given a target instance $x$, the LLM identifies perturbable factors and modifies them \emph{without changing the overall solution structure}, producing a neighborhood set
\begin{equation}
\mathcal{N}(x) = \{x^{(0)}, x^{(1)}, \dots, x^{(K)}\},
\end{equation}
where $x^{(0)}$ denotes the original instance. Perturbations follow predefined types (parameter variations, entity substitutions, condition adjustments), and labels for perturbed instances are obtained by combining LLM reasoning with white-box tools (e.g., numerical calculators).

\paragraph{Reasoning Behavior Evaluation on Perturbations.}
For each instance in $\mathcal{N}(x)$, we obtain reasoning steps and predicted answers, then quantify for each step $s_i$: (i) \emph{semantic correctness} $C_i \in \{0,1\}$, determined by an LLM semantic verifier $g(s_i, s_i^{\mathrm{ref}})$ against the reference step; and (ii) \emph{blind executability} $R_i = n_i^{\mathrm{exec}} / |\mathcal{N}(x)|$, the fraction of perturbed instances for which $s_i$ executes successfully. The step reliability weight is $W_i = C_i \cdot R_i$, jointly capturing semantic correctness and robustness to local perturbations.

\paragraph{Local Feasible-Region Graph Construction.}
We aggregate all reasoning trajectories observed across $\mathcal{N}(x)$ into a directed acyclic graph $G = (S, E)$, where each node $S_i$ is a merged reasoning step weighted by $W_i \in [0,1]$ and edges encode observed step dependencies. This graph provides a structured representation of the reasoning paths the LLM may exhibit on nearby inputs, together with their correctness and execution likelihoods, enabling local decision-behavior analysis.

\begin{wrapfigure}{r}{0.6\linewidth}
\vspace{-10pt}
\centering
\includegraphics[height=0.4\textheight]{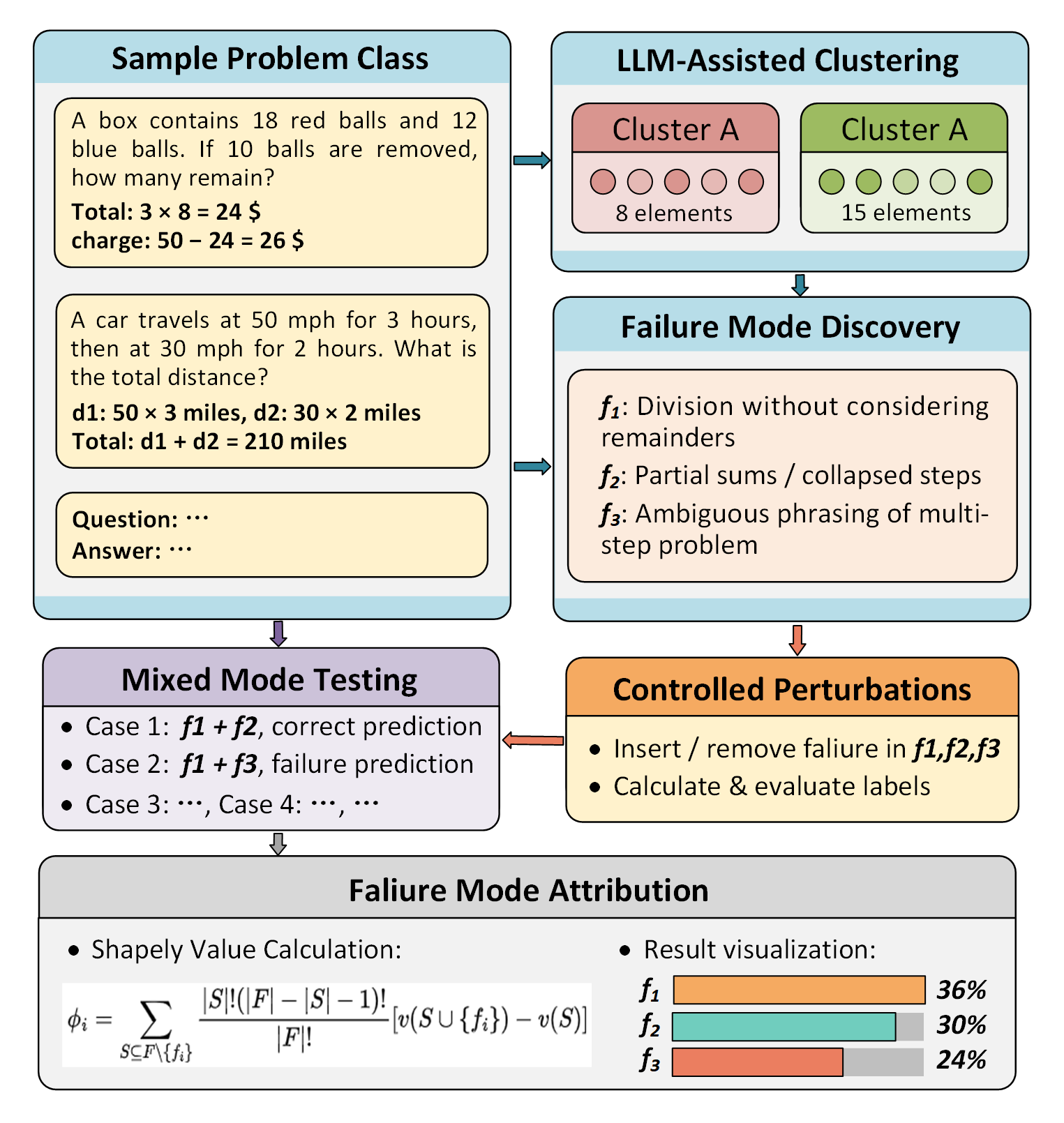}
\vspace{-10pt}
\caption{Overview of the proposed failure mode discovery and attribution framework. Given a sample problem class, semantically similar instances are grouped via LLM-assisted clustering, followed by systematic failure mode discovery. Controlled perturbations and mixed-mode testing are then performed to evaluate reasoning behavior under failure combinations. Finally, Shapley value attribution quantifies the causal impact of each failure mode on prediction outcomes. }
\vspace{-0.8cm}
\label{fig:framework}
\end{wrapfigure}

\subsection{Cluster-Level Failure Mode Analysis}
\label{sec:failure_mode}

Local feasible-region modeling captures stability within individual-sample neighborhoods but does not directly reveal class-level failure mechanisms. We therefore introduce a cluster-level failure mode analysis module that automatically discovers recurring reasoning deficiencies within a problem class and quantifies their influence on prediction outcomes (Figure~\ref{fig:framework}).

\paragraph{Failure Mode Discovery.}
We partition the dataset into small-scale clusters $C = \{x_1, \dots, x_n\}$ via LLM-assisted inductive summarization, where each cluster contains samples with comparable reasoning structures.

We define a failure mode as a minimal structural condition whose presence increases the likelihood of incorrect prediction, represented as a binary indicator $f_i : (x, E) \rightarrow \{0,1\}$. Each mode must be identifiable from the reasoning trace or input structure, manipulable via controlled perturbations, and locally minimal in influence. Failure modes are automatically extracted by an LLM-driven procedure: for incorrectly predicted samples, generated traces are aligned with reference solutions to identify divergent steps, which the LLM then abstracts into candidate failure descriptions (affected components, input fragments, hypothesized cause). Candidates are normalized and merged via semantic clustering into a compact failure mode set $F = \{f_1, \dots, f_K\}$.

\paragraph{Failure Importance Estimation.}
To quantify each mode's influence, we construct counterfactuals by injecting selected failure modes into correctly solved samples and removing/simplifying them in incorrectly solved ones; ground-truth labels are recomputed when problem parameters change. The perturbed samples form an augmented cluster $C'$ on which we measure prediction correctness under different failure configurations. We estimate the contribution of each mode via Shapley attribution:
\begin{equation}
\phi_i = \sum_{S \subseteq F \setminus \{f_i\}}
\frac{|S|!(|F|-|S|-1)!}{|F|!}
\left[v(S \cup \{f_i\}) - v(S)\right]
\label{eq:shapley_value}
\end{equation}
where $v(S)$ denotes prediction correctness under configuration $S$. The resulting $\phi_i$ values yield a cluster-level failure profile that highlights dominant error sources and provides an interpretable characterization of reasoning robustness within the problem class.

\section{Experiments}

\subsection{Executable Explanation Reliability Evaluation}

\paragraph{Setup.}
We evaluate on six benchmarks ($N{=}200$ per dataset, fixed random seed): \textbf{MATH}~\cite{hendrycks2021measuring}, \textbf{GSM8K}~\cite{cobbe2021training}, \textbf{BBH}~\cite{suzgun2023challenging}, \textbf{MMLU}~\cite{hendrycks2021measuring_mmlu}, \textbf{CRUXEval}~\cite{gu2024cruxeval}, and \textbf{HotpotQA}~\cite{yang2018hotpotqa}, spanning mathematical, logical, knowledge, code, and open-domain reasoning. We vary (i) prompting strategy (CoT, Zero-Shot-CoT, Self-Refine, Plan-and-Solve) with GPT-4o-mini fixed, and (ii) base model (7 APIs from 4 providers) with CoT fixed. GPT-4o-mini serves as the blind executor throughout to decouple execution quality from model capability.

\paragraph{Metrics.}
Let $N_{\mathrm{exec}}$, $N_{\mathrm{orig}}$, $N_{\mathrm{joint}}$, $N_{\mathrm{rec}}$ denote counts of samples correct under blind execution, original method, both, and recovery-only, respectively. We report:
\begin{equation}
EA = \frac{N_{\mathrm{exec}}}{N},\quad OA = \frac{N_{\mathrm{orig}}}{N},\quad EC = \frac{N_{\mathrm{joint}}}{N_{\mathrm{orig}}},\quad ERR = \frac{N_{\mathrm{rec}}}{N - N_{\mathrm{orig}}}.
\end{equation}
Execution Accuracy (EA) measures explanation self-sufficiency; Original Accuracy (OA) is the standard task accuracy of the prompting method; Execution Consistency (EC) quantifies what fraction of correct predictions remain verifiable under blind execution; Execution Recovery Rate (ERR) measures recovery from originally wrong predictions.

\begin{table*}[t]
\centering
\caption{Executable Explanation Evaluation across six benchmarks (\%). Base model: GPT-4o-mini, $N{=}200$ per dataset.}
\label{table3}
\small
\renewcommand{\arraystretch}{1.1}
\setlength{\tabcolsep}{3pt}
\begin{tabular}{l cccc cccc cccc}
\toprule
& \multicolumn{4}{c}{\textbf{MATH}} & \multicolumn{4}{c}{\textbf{MMLU}} & \multicolumn{4}{c}{\textbf{GSM8K}} \\
\cmidrule(lr){2-5}\cmidrule(lr){6-9}\cmidrule(lr){10-13}
\textbf{Method} & EA & OA & EC & ERR & EA & OA & EC & ERR & EA & OA & EC & ERR \\
\midrule
CoT            & 73.0 & 76.0 & 96.1 &  0.0 & 69.5 & 72.5 & 95.9 &  0.0 & 92.0 & 93.0 & 97.8 & 14.3 \\
Self-Refine    & 72.5 & 75.0 & 96.7 &  0.0 & 63.0 & 66.0 & 91.7 &  7.4 & 90.5 & 92.0 & 98.4 &  0.0 \\
Zero-Shot-CoT  & 74.5 & 78.5 & 93.6 &  4.7 & 59.0 & 69.0 & 83.3 &  4.8 & 93.5 & 95.0 & 97.9 & 10.0 \\
Plan-and-Solve & 70.0 & 73.0 & 95.2 &  1.9 & 56.0 & 72.5 & 72.4 & 12.7 & 92.5 & 93.0 & 98.4 & 14.3 \\
\midrule
\textbf{Avg.}  & 72.5 & 75.6 & 95.4 &  1.7 & 61.9 & 70.0 & 85.8 &  6.2 & 92.1 & 93.2 & 98.1 &  9.6 \\
\midrule
& \multicolumn{4}{c}{\textbf{BBH}} & \multicolumn{4}{c}{\textbf{CRUXEval}} & \multicolumn{4}{c}{\textbf{HotpotQA}} \\
\cmidrule(lr){2-5}\cmidrule(lr){6-9}\cmidrule(lr){10-13}
\textbf{Method} & EA & OA & EC & ERR & EA & OA & EC & ERR & EA & OA & EC & ERR \\
\midrule
CoT            & 68.5 & 71.5 & 95.8 &  0.0 & 46.5 & 48.0 & 96.9 & 0.0 & 14.5 & 32.0 & 42.2 & 1.5 \\
Self-Refine    & 46.5 & 60.5 & 69.4 & 11.4 & 39.0 & 42.0 & 89.3 & 2.6 &  5.0 & 15.0 & 30.0 & 0.6 \\
Zero-Shot-CoT  & 46.0 & 68.5 & 62.0 & 11.1 & 43.5 & 43.0 & 97.7 & 2.6 & 11.0 & 25.0 & 40.0 & 1.3 \\
Plan-and-Solve & 36.0 & 70.0 & 45.7 & 13.3 & 45.0 & 46.5 & 95.7 & 0.9 & 11.5 & 24.0 & 41.7 & 2.0 \\
\midrule
\textbf{Avg.}  & 49.3 & 67.6 & 68.2 &  9.0 & 43.5 & 44.9 & 94.9 & 1.5 & 10.5 & 24.0 & 38.5 & 1.4 \\
\bottomrule
\end{tabular}
\end{table*}

\paragraph{Results.}
Table~\ref{table3} reveals a clear executability hierarchy across task types. On structured tasks, EA closely tracks OA: GSM8K achieves avg.\ EC~=~98.1\% and MATH avg.\ EC~=~95.4\%, confirming that arithmetic reasoning explanations are highly self-contained. CRUXEval shows a distinctive pattern: despite moderate OA (44.9\%), EC remains 94.9\%, reflecting the inherently procedural nature of code traces. In contrast, knowledge-intensive tasks show larger EA--OA gaps; HotpotQA is the most extreme (avg.\ EA~=~10.5\% vs.\ OA~=~24.0\%, EC~=~38.5\%), as multi-hop QA requires external factual retrieval that cannot be encoded in reasoning steps alone. ERR is near zero across most datasets, indicating blind execution rarely corrects originally wrong predictions. These results establish a three-tier hierarchy: mathematical/code (EC~$>$~94\%), knowledge-intensive (EC~$\approx$~68--86\%), and open-domain QA (EC~$\approx$~39\%).

\begin{table*}[t]
\centering
\caption{Executable Explanation Evaluation across seven model APIs on four benchmarks (CoT, $N{=}200$, verifier: GPT-4o-mini). Models grouped by provider; horizontal rules separate provider families. Upper block: MATH and MMLU; lower block: BBH and CRUXEval.}
\label{tab:multi_model_combined}
\small
\renewcommand{\arraystretch}{1.1}
\setlength{\tabcolsep}{4pt}
\begin{tabular}{l l cccc cccc}
\toprule
& &
\multicolumn{4}{c}{\textbf{MATH}} &
\multicolumn{4}{c}{\textbf{MMLU}} \\
\cmidrule(lr){3-6}\cmidrule(lr){7-10}
\textbf{Provider} & \textbf{Model}
& EA & OA & EC & ERR & EA & OA & EC & ERR \\
\midrule
\multirow{3}{*}{OpenAI}
& GPT-4o-mini      & 71.5 & 74.0 & 95.9 &  1.9 & 70.0 & 72.0 & 95.8 &  3.6 \\
& GPT-4.1-nano     & 65.5 & 70.5 & 90.1 &  6.8 & 66.0 & 69.0 & 91.3 &  9.7 \\
& GPT-5-nano       & 93.0 & 97.0 & 95.9 &  0.0 & 73.5 & 75.0 & 98.0 &  0.0 \\
\midrule
Google
& Gemini-2.0-Flash & 46.0 & 48.0 & 95.8 &  0.0 & 65.0 & 74.5 & 81.2 & 17.6 \\
\midrule
\multirow{2}{*}{Alibaba}
& Qwen3.5-35B      & 98.0 & 98.5 & 99.0 & 33.3 & 80.5 & 81.5 & 98.8 &  0.0 \\
& Qwen3.5-Plus     & 98.0 & 98.5 & 99.5 &  0.0 & 83.0 & 84.5 & 98.2 &  0.0 \\
\midrule
DeepSeek
& DeepSeek-Chat    & 88.5 & 92.0 & 96.2 &  0.0 & 78.0 & 78.5 & 98.1 &  4.7 \\
\midrule
& &
\multicolumn{4}{c}{\textbf{BBH}} &
\multicolumn{4}{c}{\textbf{CRUXEval}} \\
\cmidrule(lr){3-6}\cmidrule(lr){7-10}
\textbf{Provider} & \textbf{Model}
& EA & OA & EC & ERR & EA & OA & EC & ERR \\
\midrule
\multirow{3}{*}{OpenAI}
& GPT-4o-mini      & 68.0 & 70.5 & 96.5 &  0.0 & 46.5 & 47.5 & 97.9 &  0.0 \\
& GPT-4.1-nano     & 63.5 & 67.5 & 82.2 & 24.6 & 46.5 & 49.5 & 93.9 &  0.0 \\
& GPT-5-nano       & 77.5 & 80.5 & 95.7 &  2.6 & 52.5 & 53.5 & 98.1 &  0.0 \\
\midrule
Google
& Gemini-2.0-Flash & 63.5 & 60.5 & 91.7 & 20.3 & 48.0 & 49.5 & 94.9 &  2.0 \\
\midrule
\multirow{2}{*}{Alibaba}
& Qwen3.5-35B      & 88.5 & 88.5 &100.0 &  0.0 & 53.5 & 53.5 &100.0 &  0.0 \\
& Qwen3.5-Plus     & 86.5 & 89.5 & 96.6 &  0.0 & 52.5 & 53.0 & 99.0 &  0.0 \\
\midrule
DeepSeek
& DeepSeek-Chat    & 79.5 & 80.0 & 98.8 &  2.5 & 53.0 & 53.0 &100.0 &  0.0 \\
\bottomrule
\end{tabular}
\end{table*}

Table~\ref{tab:multi_model_combined} shows that executable explanation quality consistently correlates with model capability across all four provider families. Stronger models yield higher EA: GPT-5-nano reaches 93.0\% on MATH and 77.5\% on BBH, while Qwen3.5-35B and Qwen3.5-Plus both attain 98.0\% on MATH. EC remains above 90\% for most combinations; the main exceptions---GPT-4.1-nano on BBH (82.2\%, ERR~=~24.6\%) and Gemini-2.0-Flash on MMLU (81.2\%, ERR~=~17.6\%)---suggest that weaker models occasionally produce structurally recoverable explanations despite incorrect original predictions. On CRUXEval, all models achieve EC~$>$~93.9\% regardless of capability level, confirming that code-execution reasoning produces inherently self-contained explanations. These findings validate the generality of the TRUE framework across diverse model families.

\subsection{Structural Validity Evaluation of Local Feasible Regions}

\paragraph{Setup.}
We experiment on MATH and MMLU-CF, constructing five clusters per dataset (10--20 instances each) via LLM-assisted grouping. For each cluster, an anchor instance is perturbed under three regimes (mild/moderate/aggressive) to build a feasible-region DAG; reference solutions for perturbed instances are verified with external tools. We assess two properties: (i)~\emph{success-rate prediction}: an LLM is given the DAG plus a CoT trace and asked to predict execution success; cross-entropy (CE) is compared against a same-budget baseline that uses anchor-only CoT sampling instead of perturbations; (ii)~\emph{trajectory coverage}: reasoning steps are semantically aligned to DAG nodes and the fraction of matched steps (Pret./GT) is reported across perturbation regimes.

\begin{table*}[t]
\centering
\caption{Success Rate Prediction Results on MATH and MMLU-CF.
Lower CE indicates better calibration. $\Delta$CE = Baseline CE $-$ DAG CE.}
\label{tab:success_rate_prediction}
\small
\renewcommand{\arraystretch}{1.1}
\setlength{\tabcolsep}{6pt}

\begin{tabular}{lccccc}
\toprule
\textbf{Category} & \textbf{Pert. SR} & \textbf{Test SR} 
& \textbf{DAG CE}$\downarrow$ & \textbf{Baseline CE}$\downarrow$ & $\boldsymbol{\Delta}$CE \\
\midrule

\multicolumn{6}{l}{\textbf{MATH}} \\
\midrule
Prealgebra        & 50.0\% & 87.5\% & \textbf{1.05} & 11.51 & +10.46 \\
Number Theory     & 83.3\% & 75.0\% & \textbf{0.75} & 1.39  & +0.64 \\
Intermediate Alg. & 33.3\% & 50.0\% & \textbf{0.30} & 0.75  & +0.45 \\
Counting \& Prob. & 50.0\% & 75.0\% & \textbf{1.72} & 8.69  & +6.97 \\
Precalculus       & 16.7\% & 37.5\% & \textbf{0.43} & 0.79  & +0.36 \\
\textit{Average}  & 46.7\% & 65.0\% & \textbf{0.85} & 4.63  & +3.78 \\

\midrule

\multicolumn{6}{l}{\textbf{MMLU-CF}} \\
\midrule
Classification and Rules  & 83.3\% & 65.0\% & 1.00 & \textbf{0.83} & -0.17 \\
Health and Medicine       & 100.0\% & 80.0\% & \textbf{0.53} & 4.51 & +3.98 \\
History and Politics      & 66.6\% & 75.0\% & \textbf{0.67} & 4.51 & +3.84 \\
Science and Technology    & 83.3\% & 65.0\% & 0.98 & \textbf{0.83} & -0.15 \\
Economics and Finance     & 100.0\% & 40.0\% & \textbf{1.69} & 13.82 & +12.13 \\
\textit{Average}          & 86.6\% & 65.0\% & \textbf{0.97} & 4.90 & +3.93 \\

\bottomrule
\end{tabular}
\end{table*}

\paragraph{Results.}
The DAG-based predictor substantially outperforms the anchor-only baseline on both datasets (Table~\ref{tab:success_rate_prediction}). On MATH, average CE drops from 4.63 to 0.85 ($-$81.6\%), with the largest gains in Prealgebra ($\Delta$CE~=~+10.46) and Counting \& Probability (+6.97). On MMLU-CF, average CE falls from 4.90 to 0.97, with marked improvements in Economics and Health domains. Pert.~SR closely tracks Test SR across clusters, confirming that the perturbations faithfully model the local data distribution. Trajectory coverage (Table~\ref{tab:dag_trajectory_matching}) increases monotonically with perturbation strength; under aggressive perturbations, MATH reaches 96.4\% for both Pret.\ and GT, while MMLU-CF achieves 66.6\% and 61.3\%. Together these results confirm that the feasible-region DAG captures sufficient structural information to characterize cluster-level reasoning behavior.

\begin{table*}[t]
\centering
\caption{DAG trajectory coverage results on MATH and MMLU-CF datasets.
Nodes and Edges denote the size of the feasible-region DAG.
Pret.(\%) represents the average trajectory coverage rate of perturbed instances,
while GT(\%) denotes the coverage rate of ground-truth reasoning trajectories.}
\label{tab:dag_trajectory_matching}
\small
\renewcommand{\arraystretch}{1.1}
\setlength{\tabcolsep}{6pt}

\begin{tabular}{lcccc|cccc}
\toprule

& \multicolumn{4}{c}{\textbf{MATH Dataset}}
& \multicolumn{4}{c}{\textbf{MMLU-CF Dataset}} \\

\cmidrule(lr){2-5}
\cmidrule(lr){6-9}

\textbf{Perturbation}
& \textbf{Nodes} & \textbf{Edges} & \textbf{Pret.(\%)} & \textbf{GT(\%)}
& \textbf{Nodes} & \textbf{Edges} & \textbf{Pret.(\%)} & \textbf{GT(\%)} \\

\midrule

Mild
& 9 & 35 & 51.9 & 34.8
& 18 & 29 & 50.8 & 48.3 \\

Moderate
& 10 & 35 & 74.2 & 62.9
& 23 & 30 & 58.8 & 53.1 \\

Aggressive
& 13 & 31 & 96.4 & 96.4
& 26 & 32 & 66.6 & 61.3 \\

\bottomrule
\end{tabular}
\end{table*}

\subsection{Cluster-Level Failure Mode Evaluation}

\paragraph{Setup.}
We evaluate on MATH, BBH (multiple-choice subset), and CRUXEval, constructing two representative clusters per dataset (six total) via GPT-4o-mini-assisted inductive summarization. For each cluster, five high-impact failure modes are selected and counterfactual variants are generated by (a) injecting failure conditions into correctly predicted instances and (b) removing them from incorrect ones; ground-truth labels are recomputed where needed. Shapley values $\phi_i$ are estimated from the characteristic function $v(S)$ defined by prediction correctness under each failure configuration (Eq.~\ref{eq:shapley_value}). Stability is assessed by repeating the full pipeline on subsampled clusters of size $n \in \{5, 10, 20, 40\}$, evaluating mode-set overlap (Jaccard) and ranking consistency (Kendall's $\tau$).

\begin{figure*}[t]
\centering
\includegraphics[width=\textwidth]{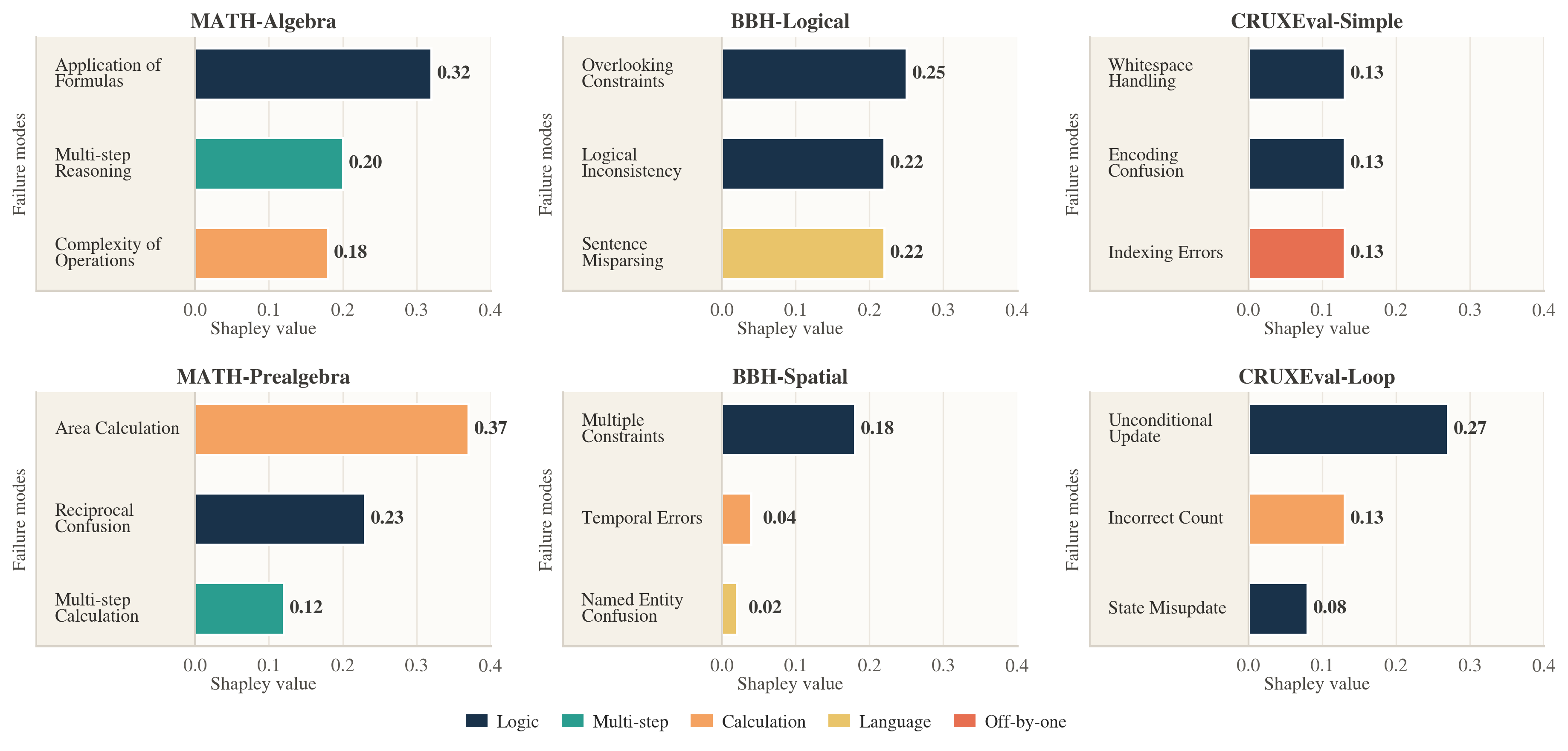}
\caption{Cluster-level failure mode profiles on MATH, BBH, and CRUXEval. Top-3 failure modes per cluster ranked by Shapley value~$\phi$. Reasoning failures concentrate in domain-specific structural patterns rather than distributing uniformly.}
\label{tab:failure_modes_big}
\end{figure*}

\paragraph{Results.}
Figure~\ref{tab:failure_modes_big} shows that reasoning failures are not uniformly distributed but concentrated in domain-specific structural patterns. On MATH, procedural errors dominate: Area Calculation ($\phi{=}0.37$) and Application of Formulas ($\phi{=}0.32$) are the top modes in Prealgebra and Algebra respectively. On BBH, constraint-management failures lead---Overlooking Constraints ($\phi{=}0.25$) and Logical Consistency Errors ($\phi{=}0.22$) in Logical Reasoning, while Spatial-Temporal shows a more diffuse distribution (max $\phi{=}0.18$). On CRUXEval, the Simple Functions cluster exhibits three equal-weight low-level errors ($\phi{=}0.13$ each), while Loop \& Recursion is dominated by Unconditional Update ($\phi{=}0.27$), reflecting difficulty in tracking mutable state. These domain-specific profiles confirm the Shapley framework's ability to pinpoint fine-grained structural weaknesses.

\begin{table*}[t]
\centering
\caption{Failure mode discovery stability under subsampling (mean$\,\pm\,$std over 3 trials). Jac: Jaccard; $\tau$: Kendall’s~$\tau$.}
\label{tab:stability}
\small
\renewcommand{\arraystretch}{1.1}
\setlength{\tabcolsep}{4pt}
\resizebox{\textwidth}{!}{%
\begin{tabular}{l c cc cc cc cc cc cc}
\toprule
& &
\multicolumn{2}{c}{\textbf{MATH-Algebra}} &
\multicolumn{2}{c}{\textbf{MATH-Prealgebra}} &
\multicolumn{2}{c}{\textbf{BBH-Logical}} &
\multicolumn{2}{c}{\textbf{BBH-Spatial}} &
\multicolumn{2}{c}{\textbf{CRUX-Simple}} &
\multicolumn{2}{c}{\textbf{CRUX-Loop}} \\
\cmidrule(lr){3-4}\cmidrule(lr){5-6}\cmidrule(lr){7-8}\cmidrule(lr){9-10}\cmidrule(lr){11-12}\cmidrule(lr){13-14}
$n$ & & Jac & $\tau$ & Jac & $\tau$ & Jac & $\tau$ & Jac & $\tau$ & Jac & $\tau$ & Jac & $\tau$ \\
\midrule
5  & & .59$_{\pm.11}$ & .22$_{\pm.88}$ & .70$_{\pm.23}$ & .22$_{\pm.57}$ & .59$_{\pm.11}$ & .67$_{\pm.27}$ & .89$_{\pm.16}$ & .49$_{\pm.13}$ & .89$_{\pm.16}$ & .18$_{\pm.14}$ & .67$_{\pm.00}$ & .00$_{\pm.27}$ \\
10 & & .70$_{\pm.23}$ & .71$_{\pm.28}$ & .59$_{\pm.11}$ & .56$_{\pm.16}$ & .70$_{\pm.23}$ & .51$_{\pm.35}$ & .89$_{\pm.16}$ & .49$_{\pm.13}$ & .89$_{\pm.16}$ & .13$_{\pm.34}$ & .78$_{\pm.16}$ & .07$_{\pm.29}$ \\
20 & & .67$_{\pm.00}$ & .67$_{\pm.27}$ & .67$_{\pm.00}$ & .67$_{\pm.00}$ & .59$_{\pm.11}$ & .67$_{\pm.27}$ & .67$_{\pm.00}$ & .56$_{\pm.16}$ & .89$_{\pm.16}$ & .44$_{\pm.26}$ & 1.00$_{\pm.00}$ & .47$_{\pm.47}$ \\
40 & & .59$_{\pm.11}$ & .44$_{\pm.42}$ & .78$_{\pm.16}$ & .44$_{\pm.31}$ & .67$_{\pm.00}$ & .33$_{\pm.00}$ & .78$_{\pm.16}$ & .36$_{\pm.03}$ & .89$_{\pm.16}$ & .33$_{\pm.25}$ & .89$_{\pm.16}$ & .38$_{\pm.33}$ \\
\bottomrule
\end{tabular}}
\end{table*}

Table~\ref{tab:stability} reports failure mode discovery stability measured by Jaccard similarity and Kendall’s $\tau$ across six clusters from three datasets as a function of subsample size.

Across all clusters, Jaccard similarity is consistently moderate to high ($\geq 0.59$), indicating that the set of discovered failure modes is robust even with small sample sizes. CRUXEval clusters exhibit the highest mode overlap stability (Jaccard $\geq 0.67$ at all sample sizes), likely because code execution errors fall into well-defined structural categories. The CRUXEval Loop \& Recursion cluster achieves perfect mode overlap (Jaccard~=~1.00) at $n{=}20$.

Ranking stability (Kendall’s $\tau$) shows more variation. MATH clusters achieve moderate ranking consistency ($\tau \approx 0.44$--$0.71$ for $n \geq 10$), while BBH Spatial-Temporal maintains stable rankings across all sizes ($\tau \approx 0.36$--$0.56$). CRUXEval clusters show lower stability at small sample sizes ($\tau \approx 0.00$--$0.18$ at $n{=}5$), but improve with larger samples ($\tau \approx 0.33$--$0.47$ at $n \geq 20$).

These results demonstrate that the proposed failure mode discovery framework achieves reliable mode identification even with limited samples ($n{=}10$--$20$), confirming its practical applicability for cluster-level reasoning analysis.

\subsection{Ablation Study and Error Analysis}
\label{sec:ablation}

\paragraph{Impact of Prompting Strategy on Faithfulness.}
We evaluate four widely-used prompting strategies—CoT~\citep{wei2022chain}, Zero-Shot-CoT~\citep{kojima2022large}, Self-Refine~\citep{madaan2023self}, and Plan-and-Solve~\citep{wang2023plan}—to assess how prompting complexity affects reasoning faithfulness as measured by the Hidden Failure Rate~(HFR~$= 1 - \text{EC}$).
Table~\ref{tab:ablation_methods} reports OA (\%) and HFR (\%) for each method on all six benchmarks.

\begin{table*}[t]
\centering
\caption{Ablation over prompting strategies: Original Accuracy (OA\%) and Hidden Failure Rate (HFR\% $= 1{-}$EC). Lower HFR indicates more faithful reasoning. Best HFR per dataset in \textbf{bold}.}
\label{tab:ablation_methods}
\small
\renewcommand{\arraystretch}{1.1}
\setlength{\tabcolsep}{3pt}
\begin{tabular}{lcccccccccccc}
\toprule
 & \multicolumn{2}{c}{\textbf{MATH}} & \multicolumn{2}{c}{\textbf{GSM8K}} & \multicolumn{2}{c}{\textbf{MMLU}} & \multicolumn{2}{c}{\textbf{BBH}} & \multicolumn{2}{c}{\textbf{CRUXEval}} & \multicolumn{2}{c}{\textbf{HotpotQA}} \\
\cmidrule(lr){2-3}\cmidrule(lr){4-5}\cmidrule(lr){6-7}\cmidrule(lr){8-9}\cmidrule(lr){10-11}\cmidrule(lr){12-13}
\textbf{Method} & OA & HFR & OA & HFR & OA & HFR & OA & HFR & OA & HFR & OA & HFR \\
\midrule
CoT              & 76.1 & \textbf{4.0}  & 93.0 & 2.2          & 72.5 & \textbf{4.1}  & 71.5 & \textbf{4.2}  & 48.0 & 3.1          & 32.0 & 57.8 \\
Zero-Shot-CoT    & 78.5 & 6.4           & 95.0 & 2.1          & 69.0 & 16.7          & 68.5 & 38.0          & 43.0 & \textbf{2.3} & 25.0 & 60.0 \\
Self-Refine      & 75.2 & 3.3           & 92.0 & \textbf{1.6} & 66.0 & 8.3           & 60.5 & 30.6          & 42.0 & 10.7         & 15.0 & 70.0 \\
Plan-and-Solve   & 73.0 & 4.8           & 93.0 & \textbf{1.6} & 72.5 & 27.6          & 70.0 & 54.3          & 46.5 & 4.3          & 24.0 & \textbf{58.3} \\
\bottomrule
\end{tabular}
\end{table*}

Two key findings emerge. First, \emph{more elaborate prompting does not guarantee more faithful reasoning.} Plan-and-Solve achieves similar OA to CoT on BBH ($70.0\%$ vs.\@ $71.5\%$) but incurs a dramatically higher HFR ($54.3\%$ vs.\@ $4.2\%$): over half of its apparently correct solutions fail blind execution verification. Similarly, Zero-Shot-CoT on MMLU reaches $69.0\%$ OA while producing $16.7\%$ hidden failures. Second, \emph{faithfulness degrades sharply with task complexity.} On structured-output tasks (MATH, GSM8K), all methods achieve HFR~$\leq 6.4\%$, while on open-ended multi-hop QA (HotpotQA) HFR exceeds $57\%$ for every method—indicating that the model's stated reasoning rarely fully encodes its answer derivation for complex QA.

\begin{figure}[t]
\centering
\includegraphics[width=\columnwidth]{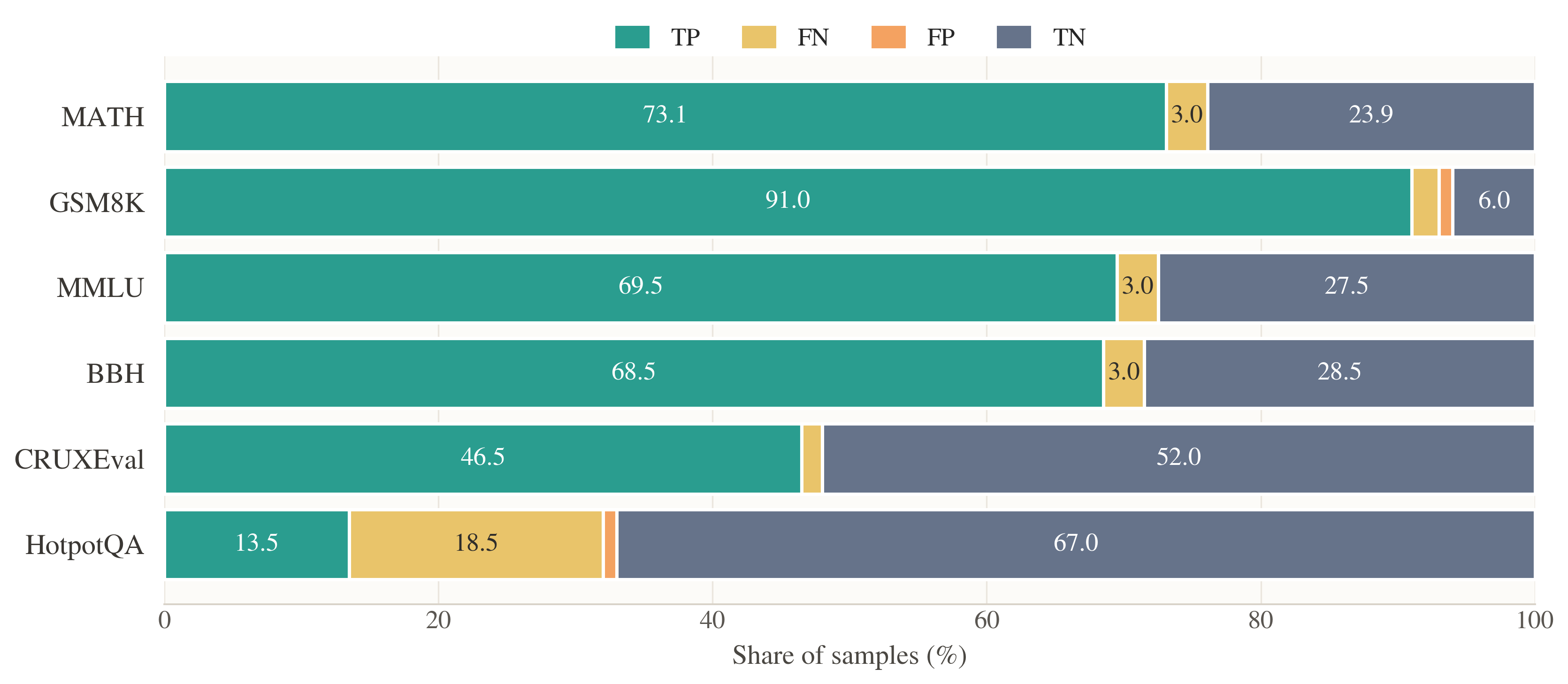}
\caption{Error breakdown (\% of $n{=}200$ samples) for CoT. FP~$\approx 0$ across all tasks; FN (hidden failures) dominates on HotpotQA.}
\label{tab:error_breakdown}
\end{figure}

\paragraph{Error Breakdown Analysis.}
Figure~\ref{tab:error_breakdown} details the four-way error decomposition for CoT across all datasets.
\emph{True Positive (TP)}: both CoT answer and blind-executed answer are correct;
\emph{False Negative (FN)}: CoT answer correct but execution fails (hidden failure);
\emph{False Positive (FP)}: execution succeeds but CoT answer is wrong (rare, indicates lucky execution);
\emph{True Negative (TN)}: both fail.

The FP rate is near zero across all benchmarks, confirming that blind execution does not fabricate correct answers and the verifier is conservative. The dominant failure mode is FN (hidden failures), which is negligible on arithmetic tasks (MATH: 3.0\%, GSM8K: 2.0\%) but substantial on HotpotQA (18.5\%). The high TN rate on CRUXEval (52.0\%) and HotpotQA (67.0\%) reflects the inherent difficulty of those tasks rather than a failure of the TRUE framework itself. Together these results confirm the central claim of TRUE: OA systematically overestimates reasoning reliability, and EC (or equivalently HFR) provides an executable audit of faithfulness that OA cannot capture.

\section{Conclusion}

We propose TRUE, a unified framework for analyzing LLM reasoning through executable verification and perturbation-driven structural analysis. Its three components---executable explanations with blind-execution verification, perturbation-guided feasible-region DAGs, and Shapley-based cluster-level failure mode analysis---jointly characterize reasoning behavior at instance, local, and class levels. Experiments on multiple reasoning benchmarks confirm that executable explanations reliably encode solution procedures, feasible-region DAGs improve execution success prediction, and cluster-level analysis identifies stable, interpretable failure modes even under limited samples. Together, these results establish a principled paradigm for interpretable and verifiable LLM reasoning analysis.

\paragraph{Limitations.}
Two limitations warrant attention. First, the blind execution verifier $V$ relies on an LLM for step interpretation, bounding its reliability to the quality of the underlying model. Second, exact Shapley attribution over all $2^K$ subsets limits the framework to small $K$ (we use $K{=}5$); efficient Shapley approximation algorithms would be needed to scale to larger mode sets.

\bibliographystyle{unsrtnat}
\bibliography{reference}

\end{document}